\documentclass{llncs}
\usepackage{llncsdoc}
\usepackage{graphicx}
\usepackage{amsfonts}
\usepackage{amsmath}
\usepackage{algorithm}
\usepackage{algpseudocode}
\usepackage[caption=false]{subfig}
\usepackage{array}
\usepackage{multirow}

\usepackage{geometry}
\begin{document}
\title{A novel method for automatic localization of joint area on knee plain radiographs}

\author{Aleksei Tiulpin\inst{1} \and Jerome Thevenot\inst{1} \and Esa Rahtu \inst{2} \and Simo Saarakkala\inst{1,3}}

\institute{Research Unit of Medical Imaging, Physics and Technology, \\University of Oulu, Finland
\and
Center for Machine Vision and Signal Analysis,\\ University of Oulu, Finland
\and
Department of Diagnostic Radiology, Oulu University Hospital, Finland}

\maketitle
\begin{abstract}
Osteoarthritis (OA) is a common musculoskeletal condition typically diagnosed from radiographic assessment after clinical examination. However, a visual evaluation made by a practitioner suffers from subjectivity and is highly dependent on the experience. Computer-aided diagnostics (CAD) could improve the objectivity of knee radiographic examination. The first essential step of knee OA CAD is to automatically localize the joint area. However, according to the literature this task itself remains challenging. The aim of this study was to develop novel and computationally efficient method to tackle the issue.
Here, three different datasets of knee radiographs were used (n = 473/93/77) to validate the overall performance of the method. Our pipeline consists of two parts: anatomically-based joint area proposal and their evaluation using Histogram of Oriented Gradients  and the pre-trained Support Vector Machine  classifier scores. 
The obtained results for the used datasets show the mean intersection over the union equals to: 0.84, 0.79 and 0.78. Using a high-end computer, the method allows to automatically annotate conventional knee radiographs  within 14-16ms and high resolution ones within 170ms. Our results demonstrate that the developed method is suitable for large-scale analyses.

\end{abstract}
\section{Introduction}
Hip and knee osteoarthritis (OA) are globally ranked as the 11th highest contributor to disability \cite{cross2014global}. Typically, OA is diagnosed at a late stage when there is no cure available anymore and when joint replacement surgery becomes the only remaining option. However, if the disease could be diagnosed at an early stage, its progression could be slowed down or even stopped. Such diagnostics is currently possible; however, it involves the use of expensive imaging modalities, which is clinically unfeasible in the primary health care.

X-ray imaging is the most popular and cheap method for knee OA diagnostics \cite{demehri2015conventional,cibere2006we}. However, the diagnostics at an early stage based on this modality has still some limitations due to several factors. A visual evaluation made by a practitioner suffers from subjectivity and is highly dependent on the experience. Eventually it has been reported that radiologists misdiagnosed OA in 30\% of the cases, and that in 20\% of the cases a specialist disagrees with his/her own previous decision after a period of time \cite{berlin1996malpractice,pitman2006perceptual}. Therefore, to make the diagnostic process more systematic and reliable, computer-aided diagnostics (CAD) can be used to reduce the impact of the subjective factors which alter the diagnostics. 

In the literature, multiple attempts have been made to approach knee OA CAD from X-ray images \cite{Antony16,thomson2015automated,stachowiak2016detection,podsiadlo2014trabecular,shamir2009early}. These studies indicate existence of two parts in the diagnostic pipeline: the region of interest (ROI) localization and, subsequently, a classification -- OA diagnostics from localized ROI(s). It has been reported (see, section below), that this problem remains challenging and requires a better solution. 

The aim of our study was to propose a novel and efficient knee joint area localization algorithm, which is applicable for large-scale knee X-ray studies. Our study has the following novelties:
\begin{itemize}
\item We propose a new approach to generate and score knee joint area proposals.
\item We report a cross-dataset validation, showing that the our method performs similarly for three different datasets and drastically outperforms the baseline.
\item Finally, we show, that the developed method can be used to accurately annotate from hundreds of thousands to millions of X-ray images per day.
\end{itemize}

\section{Related work}\label{related_work}

In the literature, multiple approaches have been used to localize ROIs within radiographs: manual \cite{hirvasniemi2014quantification,woloszynski2010signature}, semi-automatic \cite{duryea2000trainable} and in a fully automatic manner \cite{Antony16,podsiadlo2008automated,shamir2009early}. To the best of our knowledge, only the studies by Anthony \textit{et al.} \cite{Antony16}  and Shamir \textit{et al.} \cite{shamir2009early} focused on knee joint area localization.
While the problem of knee joint area localization can be implicitly solved by annotating the anatomical bone landmarks using deformable models \cite{seise2005double,lindner2013development}, it would be unfeasible to perform large-scale studies since the use of these algorithms is computationally costly. Currently, despite the presence of multiple approaches, no studies have reported so far their applicability among different datasets. This cross-dataset validation is crucial for the development of clinically applicable solutions.

In the recently published large scale study \cite{Antony16}, two methods were analyzed: a template matching adapted from a previously published work \cite{shamir2009early} and a sliding-window approach. Both methods showed limited localization ability, however, the sliding-window approach demonstrated a better performance. In particular, this approach was designed to find knee joint centers in radiographic images using Sobel \cite{duda1973pattern} gradients and a linear Support Vector Machine (SVM) classifier. For each sliding window, an SVM score was computed and eventually the patch having the best score was selected. Subsequently, a $300\times300$ pixels region was drawn around the selected patch center. As the localization metric, the authors used intersection over the union (IoU), which is also called the Jaccard index, between the drawn region and the manual annotation:
\begin{equation}
IoU=\frac{A\cap B}{A\cup B},
\end{equation}
where $A$ is the manual annotation and $B$ is  the detected bounding box.
While the sliding window approach was better than the template matching (mean IoU over the dataset was 0.36 vs. 0.1), the performance was still insufficient to perform further OA CAD, as it was indicated by the authors themselves.  Consequently, there is a need for more effective methods for ROI localization since nowadays large scale studies are possible due to the availability of multiple image cohorts like Osteoarthritis Initiative (OAI) \cite{eckstein2006double} and Multicenter Osteoarthritis study (MOST) \cite{englund2009meniscal}. These cohorts contain follow-up data of thousands of normal subjects and subjects with knee, hip and wrist OA.

Finally, it should be mentioned, that the problem of joint area localization is not limited to only knee radiographs. The attention of the research community is also focused on hand radiographs, where OA occurs as well. It has been recently shown how the anatomic structure of the image can be utilized to annotate hand radiographic images \cite{huo2013automatic}. However, to the best of authors' knowledge, there are no studies in the knee domain where such information is used to segment or annotate the image. In this study, we show how such information can be used to annotate knee radiographs.

\section{Method}
\begin{figure}
\centering
       \includegraphics[width=\textwidth]{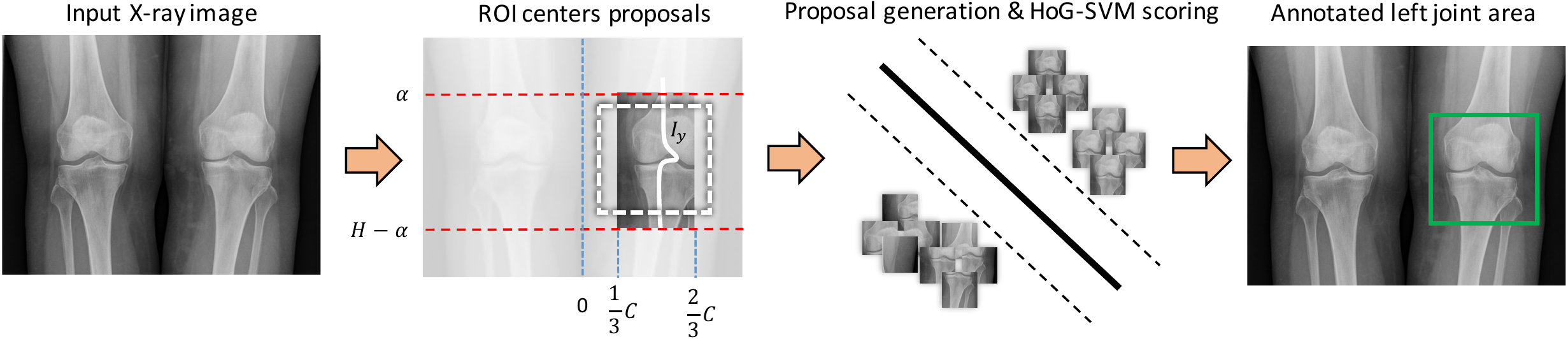}
		\caption{Schematic representation of the proposed knee joint area localization method. For more details of the method, see the sections \ref{sec:proposal} and \ref{sec:propscor}.}\label{img:propgen}
\end{figure}

In this section, we describe our joint area localization approach. The method consists of two main blocks: proposal generation and SVM-based scoring. First, we describe a limb anatomy-based proposal generation algorithm. As shown in previous object detection studies \cite{uijlings2013selective,zitnick2014edge}, object proposal approaches can significantly reduce the amount of candidate regions: from hundreds of thousands to a few thousands. Subsequently, we show the proposal scoring step, based on Histogram of Oriented Gradients (HoG)  \cite{dalal2005histograms} and SVM classifier. Schematically, our approach is presented in Figure \ref{img:propgen}.

\subsection{Data pre-processing}
To approach the joint area localization problem on knee radiographs, it would be possible to utilize an exhaustive sliding window approach, however, it is time consuming and inadequate for big data analysis. Thus, we propose an anatomically-based proposal generation approach to significantly reduce the amount of checked joint locations. Our method is trained for only one leg, despite the fact that two legs are usually imaged (see the leftmost block in Figure  \ref{img:propgen}). We obtain the method working for both legs by mirroring the leg which is not used for training the method. Therefore, below we describe the problem for the image containing only one leg and denote it as $\mathbf{I}$ having size $C\times H$. 

\subsection{Region proposal}\label{sec:proposal}
The core idea and novelty behind our proposal generation are to utilize the anatomic structure of the limb. As the prior information of the knee joint anatomy is known, it can be efficiently utilized. Considering marginal distribution of intensities of a leg, it can be noticed, that around the joint area, there is an intensity increase due to the presence of the patella, and then a sharp intensity drop due to the space between the joints. In our approach, we detect these changes and use their locations as Y-coordinates of region proposals.

At first, we take the middle of an input image, and then sum all the values horizontally to obtain the sequence $I_y$, which corresponds to a vertical marginal distribution of pixel intensities:
\begin{equation}\label{eq:first}
I_y[i-\alpha] = \sum_{j=\frac{1}{3}C}^{\frac{2}{3}C-1}\mathbf{I}[i, j],\,\forall i\in [\alpha,H-\alpha).
\end{equation}

Here, we do not sum all the values -- instead, we use a margin $\alpha$ to ignore the outliers which are usually present in the top and the bottom of the image. This also reduces the computational complexity of the method. The next step is to identify the intensity peaks located near the patella. We apply the derivation to the obtained sequence in order to detect the anatomical changes and a moving average as a convolution with a window function $w[\cdot]$ of size $s_w$ to reduce the number of peaks. Eventually, we use a sequence obtained by taking the absolute of each of the $I_y$ values:
\begin{equation}\label{eq:second}
I_y[i] = |(I_y' * w)[i]|, \forall i\in [0, H-2\alpha).
\end{equation}

\begin{figure}
\centering 

\begin{tabular}{cc}
\subfloat[]{\includegraphics[width=0.46\textwidth]{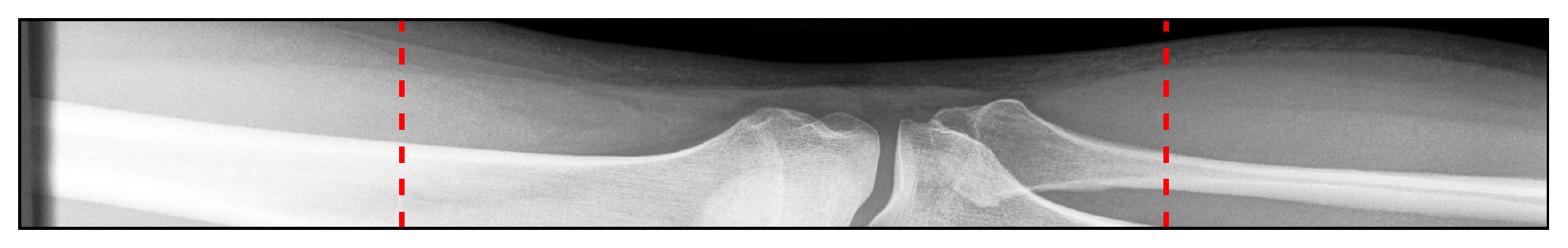}} & 
\subfloat[]{\includegraphics[width=0.46\textwidth]{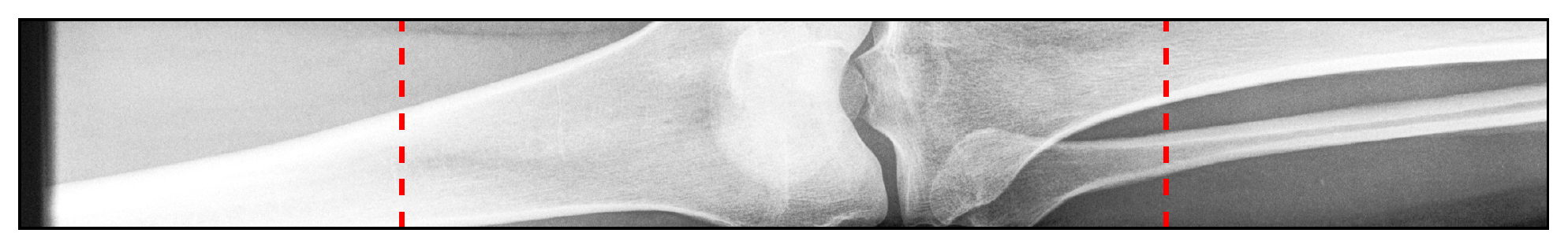}} \\
\subfloat[]{\includegraphics[width=0.46\textwidth]{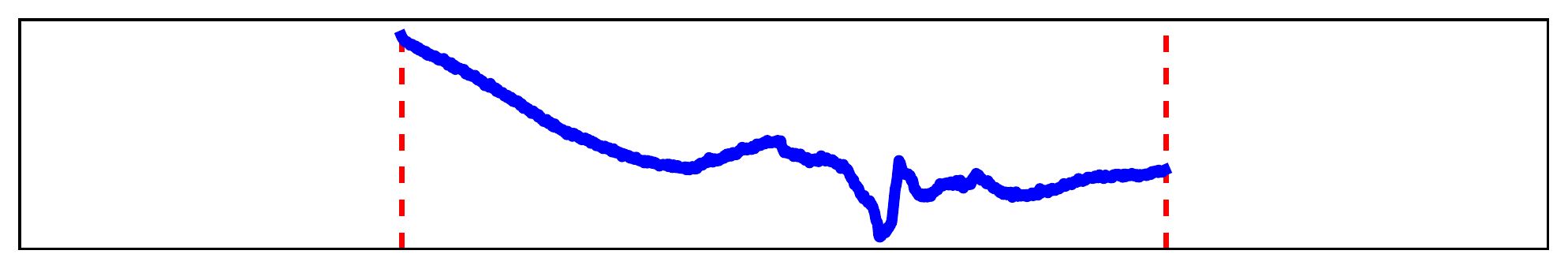}} & 
\subfloat[]{\includegraphics[width=0.46\textwidth]{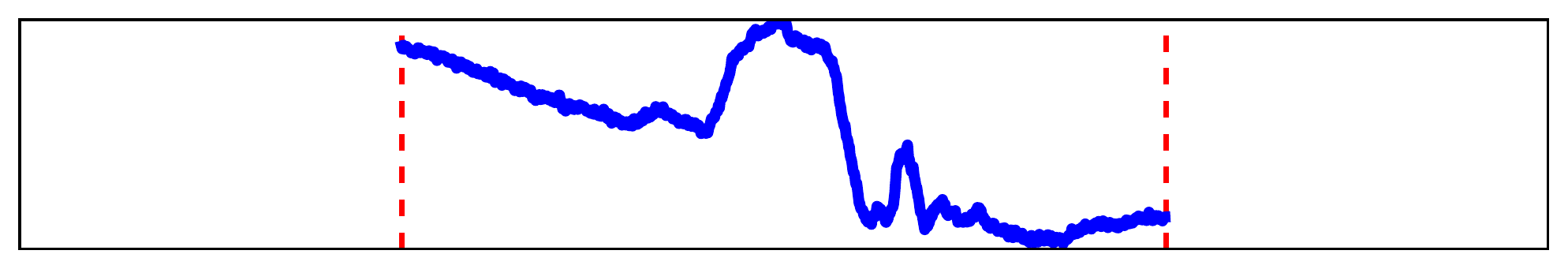}}\\
\subfloat[]{\includegraphics[width=0.46\textwidth]{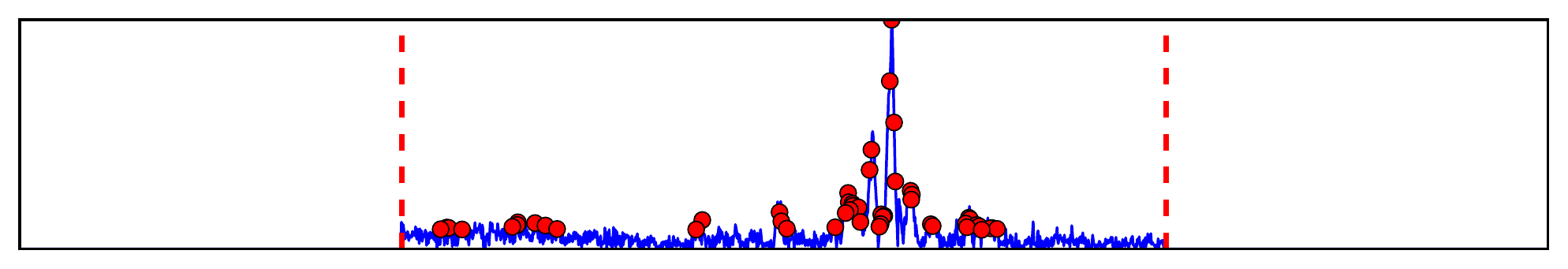}} & 
\subfloat[]{\includegraphics[width=0.46\textwidth]{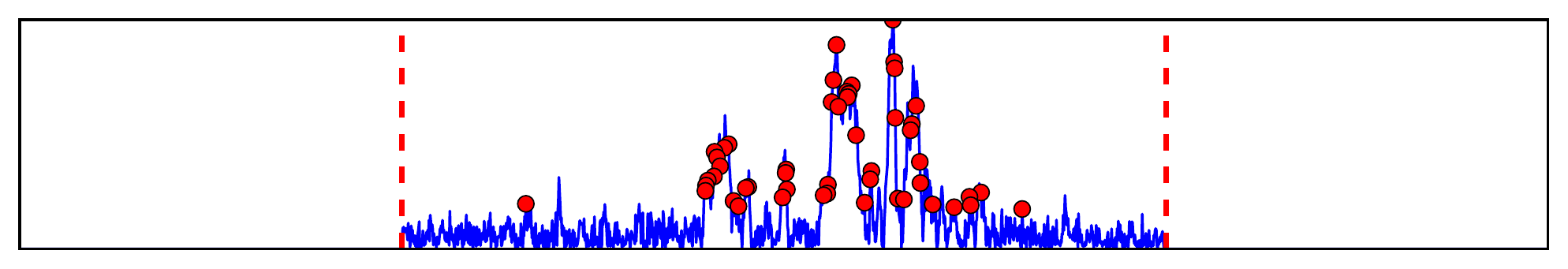}}\\
\end{tabular}
\caption{Visualization of the joint center Y-coordinate proposal procedure for the left and right leg images -- left and right columns, correspondingly. a) and b) show the sub-areas of the leg images, the red lines indicate the zones which are used for the analysis in equation \ref{eq:first}. c) and d) show the obtained sequences $I_y$ from equation \ref{eq:first}. e) and f) show the result of applying equation \ref{eq:second}, red dots indicate the values used to obtain Y-coordinates.}
\label{img:vis_y_coord}
\end{figure}

At the last step, we obtain each $k$-th index out of the top $\tau$\% of $I_y$ values. It should be mentioned, that since the margin $\alpha$ was used, it has to be added to each selected index of $I_y$. A visualization of the procedure described above is given in Figure \ref{img:vis_y_coord}.

Finally, we take the proposed Y-coordinates and as X-coordinates derive as $x=\frac{1}{2}W+j,\forall j\in [d_1,d_1+p,d_1+2p,\dots,d_2]$, where $p$ is a displacement step, which can be estimated using a validation set (see, section \ref{sec:details}). The procedure described above is repeated for each leg.

To generate the proposals at multiple scales, we use a data-driven approach, which requires a training set having manual annotations. Here, we consider a joint area to be within a square of size $S_n$, where $S_n$ is proportional to an image height $H$ with some factor $\frac{1}{Z_n}$.  Using manual annotations of training data we estimate a set  $\mathbf{Z}$ having scales $Z_n$ for each training image. Eventually, having an image $\mathbf{I}$ of size $C\times H$, we use a set $\mathbf{S}$ of proposal sizes based on the following estimations: $ S_n=\frac{1}{Z_n}H, \forall Z_n\in [\dots, \overline{\mathbf{Z}}-\sigma(\mathbf{Z}),\overline{\mathbf{Z}}, \overline{\mathbf{Z}}+\sigma(\mathbf{Z}),\dots]$, where $ \overline{\mathbf{Z}}$ is the mean and $\sigma(\mathbf{Z})$ is the standard deviation of the scales set estimated from the training data. 

To conclude, the exact amount of generated proposals  for each image is calculated as
\begin{equation}
|\mathbf{S}|[R + (R \bmod k \not = 0)]\cdot{[d_2-d_1 + ((d_2-d_1) \bmod p \not = 0)]},
\end{equation}
where $R=0.01\tau\cdot(H-2\alpha)$ . Here, we use a product rule and consider that all proposals belong to the image area. To estimate the number of proposals for the whole image with two leg image, their amount has to be multiplied by 2.

\subsection{Proposal scoring}\label{sec:propscor}
The next step in our pipeline is to train a classification algorithm in order to select the best candidate region among generated proposals. For that, we use HoG feature descriptor (to compactly describe the knee joint shape) and linear SVM classifier. This combination has been successfully used for human detection and has demonstrated excellent detection performance\cite{dalal2005histograms}. Additionally, the extraction of HoG features and SVM-based scoring are relatively fast pipeline blocks, which  significantly supported our choice of the proposal scoring approach.

The HoG-SVM pipeline is implemented as follows: for each half of the image we generate the proposals and downscale each of them into a patch of  $64\times64$ pixels. Then, we compute HoG features using the following parameters: 16 blocks with 50\% overlapping, 4 cells in a block and 9 bins \cite{dalal2005histograms}. Finally, we classify vectors of 1,764 values (7 blocks vertically times 7 blocks horizontally times 4 cells times 9 bins). Basic maximum search is used to find the best scored proposal.

\section{Experiments}
\subsection{Data}
In this study we used several data sources: 1,574 high resolution knee radiographs from MOST cohort\cite{englund2009meniscal}, 93 radiographs from \cite{multanen2015bone} (Dataset obtained from Central Finland Central Hospital, Jyv\"askyl\"a. The dataset is called "Jyv\"askyl\"a dataset" in this paper), and 77 from OKOA dataset \cite{podlipska2016comparison}. The images in all datasets contain knees with different severity of OA as well as implants. The images from MOST were used to create training, validation and test sets, while the remaining were only used to assess the generalization ability of the developed algorithm. More detailed information about the data and their usage is presented in Table \ref{tab:datasets}.

\begin{table}
\centering
\caption{Description of datasets used in this study. The training data were used to train the algorithm, validation data to find hyperparameters for the algorithm, and the testing set to assess the localization and detection performance of the proposed method.}\label{tab:datasets}
\begin{tabular}{p{2cm}p{2cm}p{2cm}p{2cm}p{3cm}}
\hline
\noalign{\smallskip}
Dataset & Training set  & Validation set & Test set & Average image size \\
\noalign{\smallskip}
\hline
\noalign{\smallskip}
MOST & 991 & 110 & 473& $3588\times4279$ \\ 
Jyv\"askyl\"a & -  & -  & 93&$2494\times2048$\\
OKOA & -  & -& 77& $2671\times2928$\\ 
\noalign{\smallskip}
\hline
\noalign{\smallskip}
\end{tabular}
\end{table}

We converted the original X-ray data from 16 bit DICOM files to a 8 bit format for standardization: we truncated their histograms between the 5th and the 99th percentiles to remove corrupted pixels. Then we normalized the images by scaling their dynamic range between 0 and 255. After that, we manually annotated all used images using ad-hoc MATLAB-based tool. Main criterion for the annotation was, that the ROI should include the joint itself and the fibula bone.

\begin{figure}
     \subfloat[Positive \label{img:pos}]{%
       \includegraphics[width=0.45\textwidth]{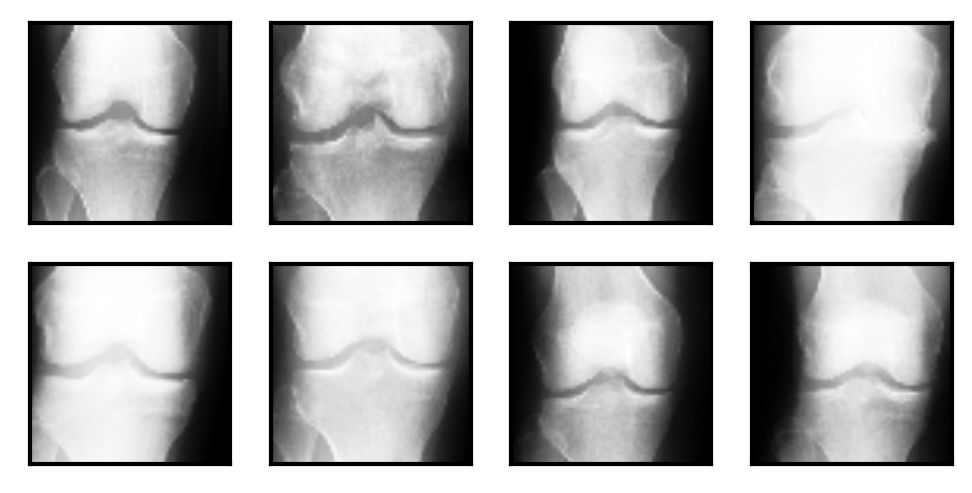}
       }
     \hfill
     \subfloat[Negative \label{img:neg}]{%
       \includegraphics[width=0.45\textwidth]{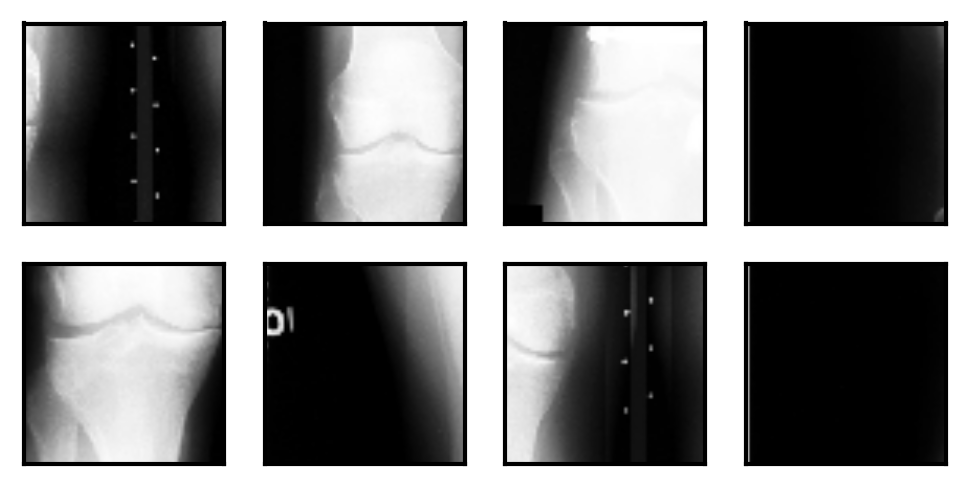}
     }
     \caption{Examples of positive and negative train patches. These examples were generated automatically using the proposal generation algorithm and manual annotations.}
     \label{img:posneg}
\end{figure}

To create a dataset for training a HoG-SVM proposal scoring, we processed the original training data as follows: for each knee on the training images, we generated the proposals and  marked them as positive if the IoU of the  manual annotation and the generated proposal was greater or equal than 0.8. Such a strict threshold was selected to make positive proposals be as close to manual annotations as possible. Examples of positive and negative proposals are given in Figure \ref{img:posneg}.

To augment the training set for more robust training, we performed the following transformations: the amount of positive was increased by a factor of 5 using rotations in the range [-2,2] degrees with a step 0.8. Here, we trained the classifier for only one leg, since the legs on the image are symmetrical. At the classification step, we used flipped proposals of the left leg image before extracting HoG features.

\subsection{Implementation details}\label{sec:details}
To extract HoG, we used OpenCV \cite{itseez2015opencv} implementation and for SVM traing we used a dual-form implementation from  LIBLINEAR \cite{REF08a} package. To find the appropriate regularization constant $C_{s}$ for SVM scorer, we used a validation set described earlier. We tried to scale the data before SVM to improve the classification results as well as hard-negative mining. However, neither of these approaches did not provided any improvement. Eventually, we found that $C_{s}=0.01$ without data scaling and hard-negative mining gives the best precision and recall on the validation set.

In our pipeline, we fixed the smoothing window width $s_w=11$ pixels, the displacement range to be $[d_1=-\frac{1}{4}C,d_2=\frac{1}{4}C]$ pixels, $k=10$ and $\tau=10$. Based on the manual annotations of the training data we estimated the set of scales $\mathbf{Z}=[3.0, 3.2, 3.4, 3.6, 3.8, 4.0, 5.0]$. After that, using the same validation set as for SVM parameters tuning, we adjusted the step $p$. Our main criterion was to find the best IoU having fast computational time. We found $p=95$ pixels (IoU=0.843, time per image -- 162ms).

\subsection{Localization performance evaluation}
\begin{figure}[h!]
     \subfloat[MOST dataset \label{img:propmost}]{%
       \includegraphics[width=0.32\textwidth]{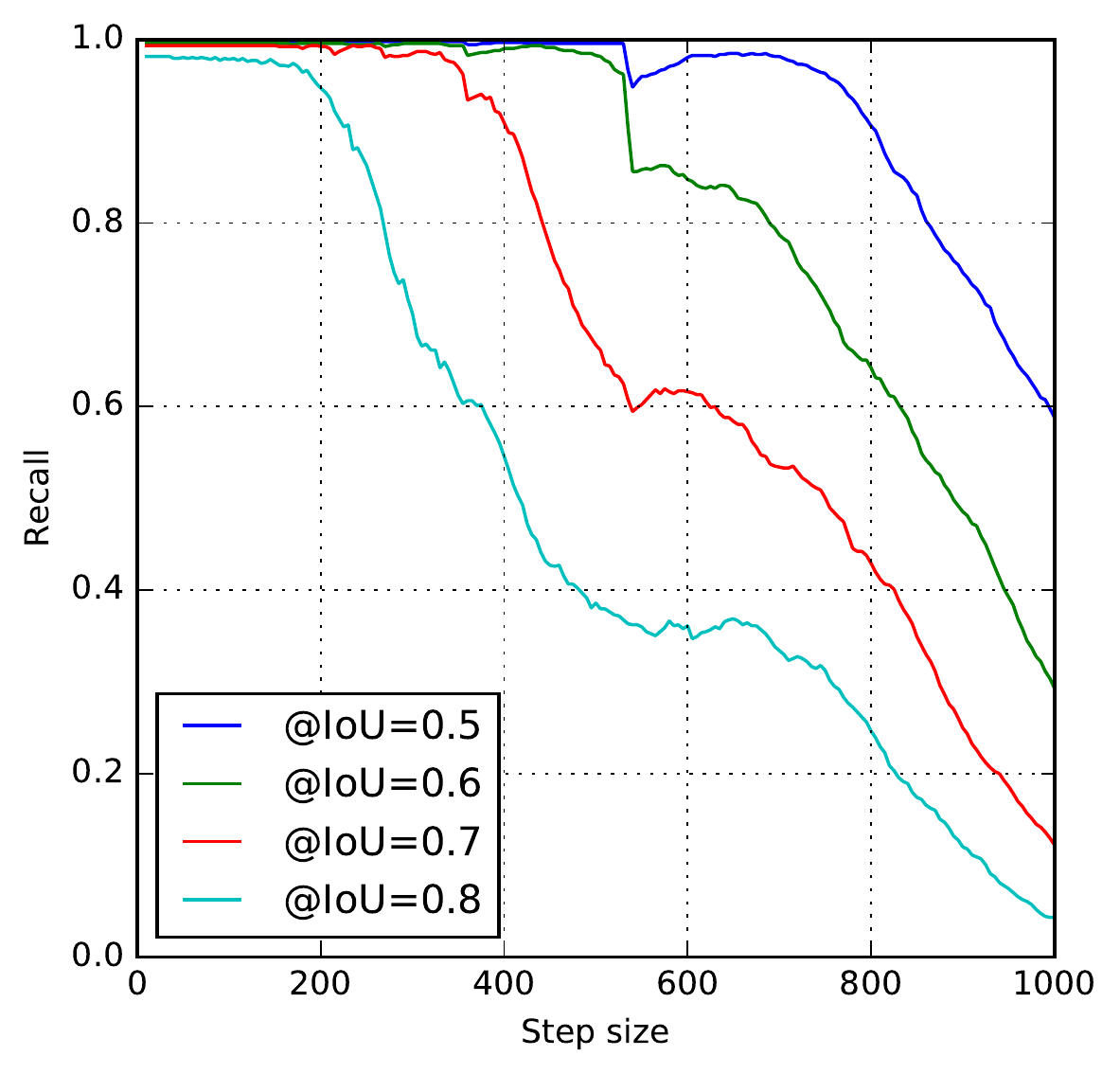}
       }
     \hfill
     \subfloat[Jyv\"askyl\"a dataset \label{img:propjkl}]{%
       \includegraphics[width=0.32\textwidth]{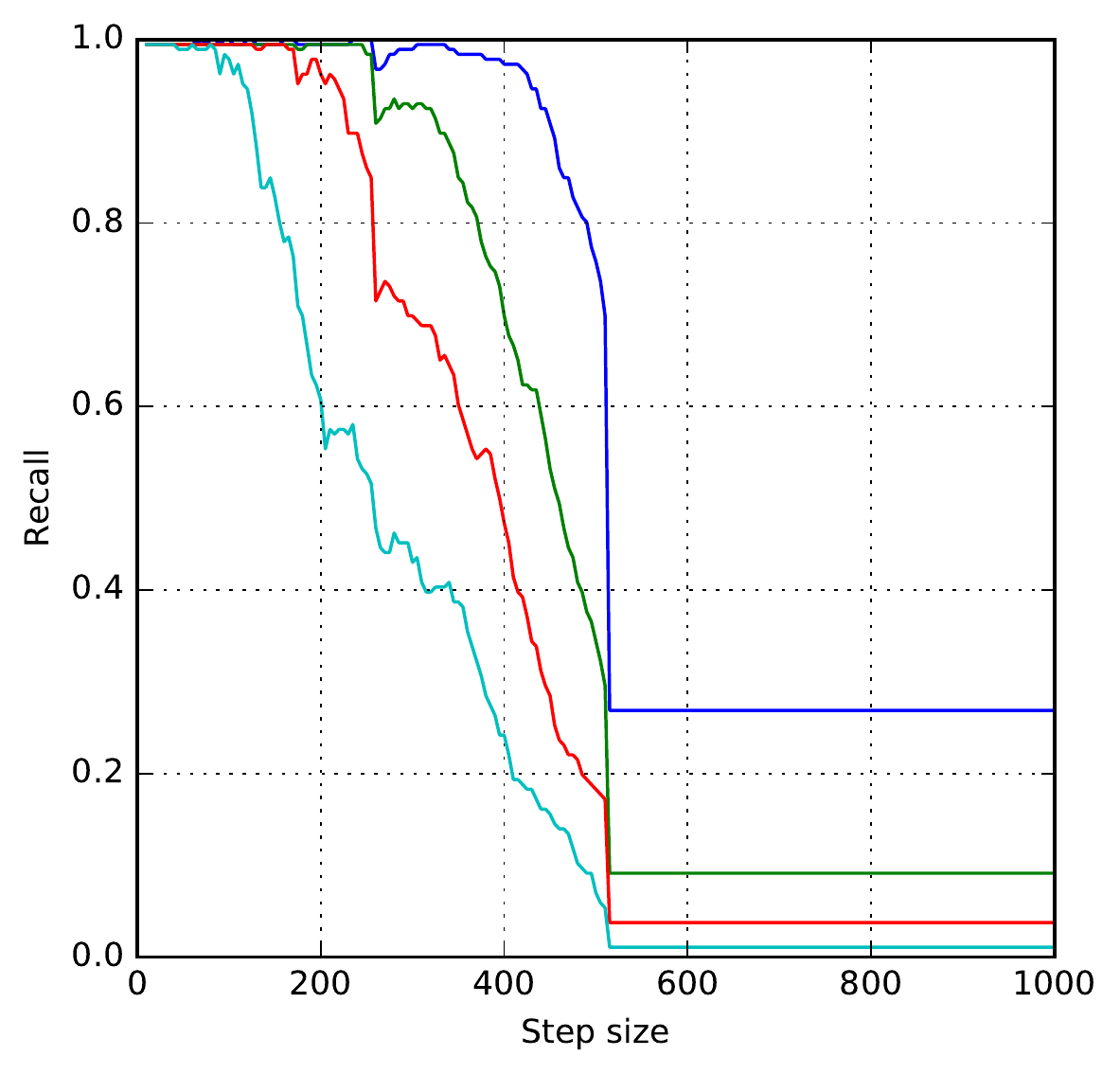}
       }
     \hfill
     \subfloat[OKOA dataset \label{img:propokoa}]{%
       \includegraphics[width=0.32\textwidth]{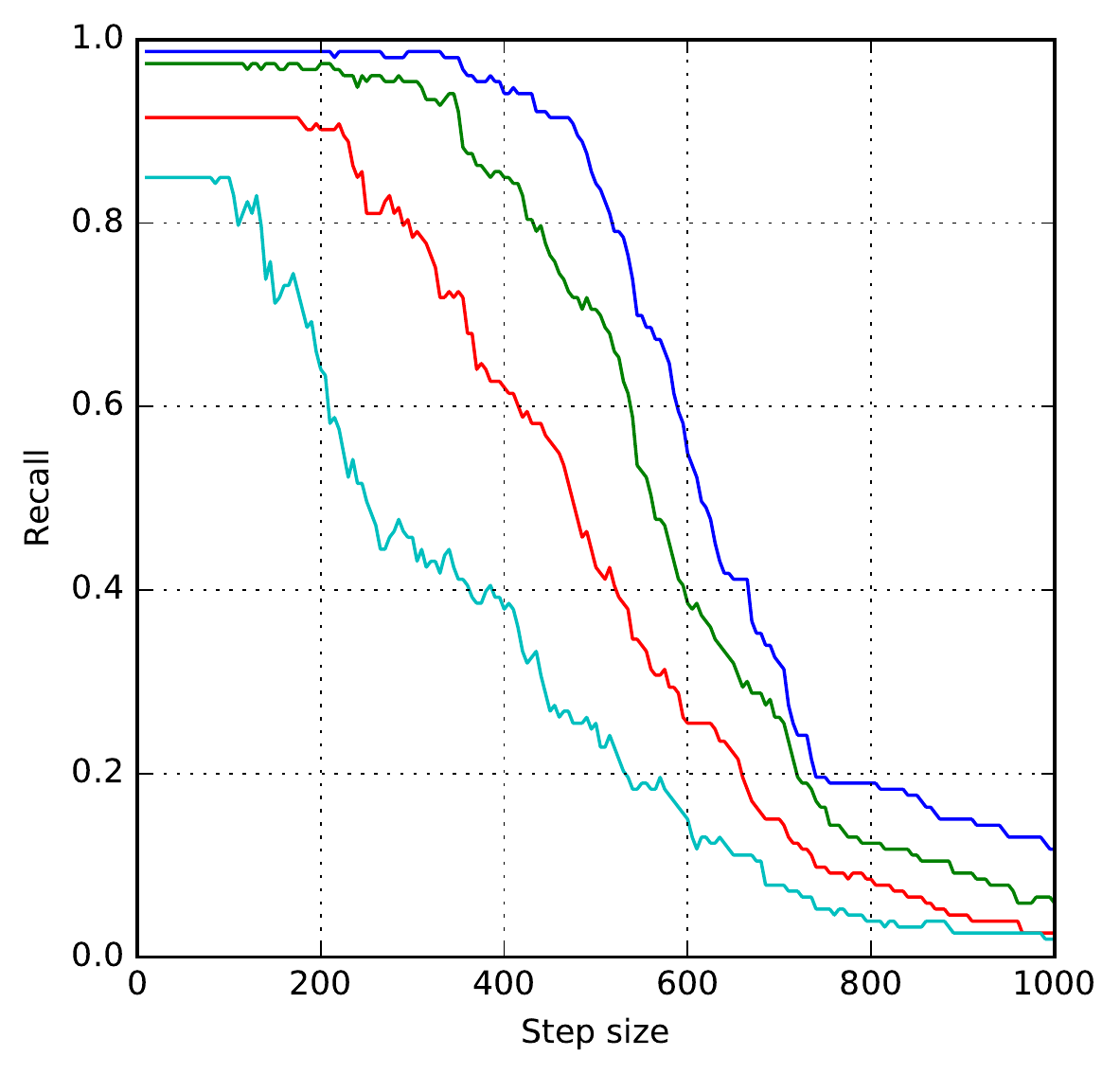}
       }
     \caption{Proposals quality evaluation for each analyzed dataset -- recall depending on the step value $p$ and different IoU thresholds. Analysis shows, that it is feasible to reach a recall above 80\% in every analyzed dataset having IoU threshold 0.8.}
     \label{img:propquality}
\end{figure}

Before performing the evaluation of our pipeline on the testing datasets, we estimated the quality of the generated proposals for each of them. The evaluations are presented in Figure \ref{img:propquality}. 

\begin{figure}
     \subfloat[]{%
       \includegraphics[width=0.3\textwidth]{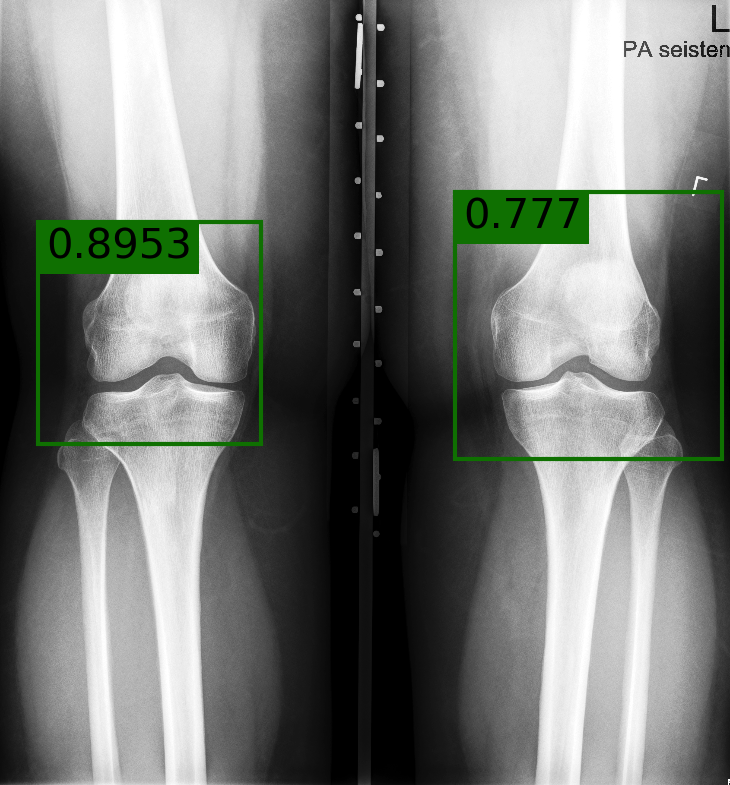}
       }
            \hfill
     \subfloat[]{%
       \includegraphics[width=0.3\textwidth]{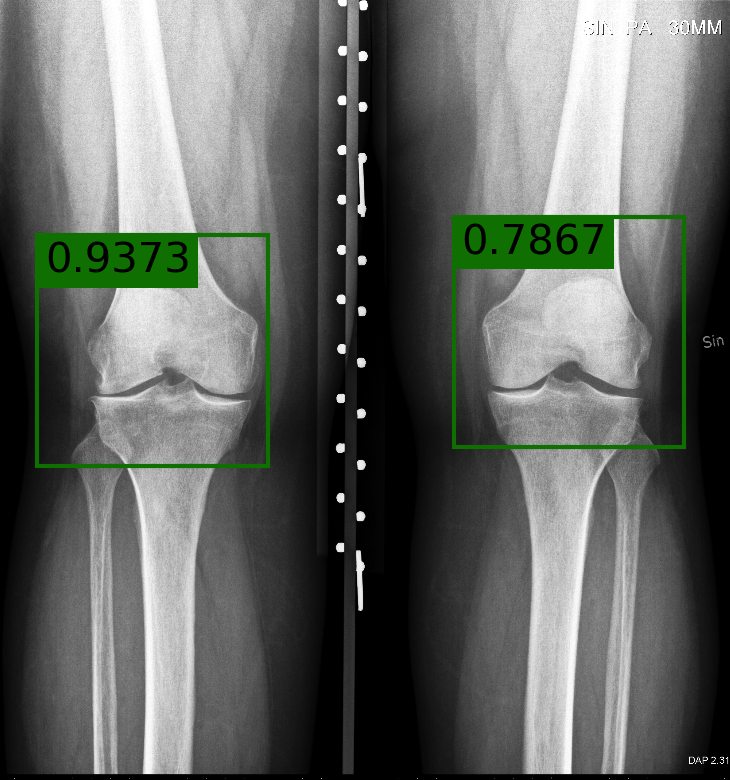}
       }
                \hfill   
        \subfloat[]{%
       \includegraphics[width=0.3\textwidth]{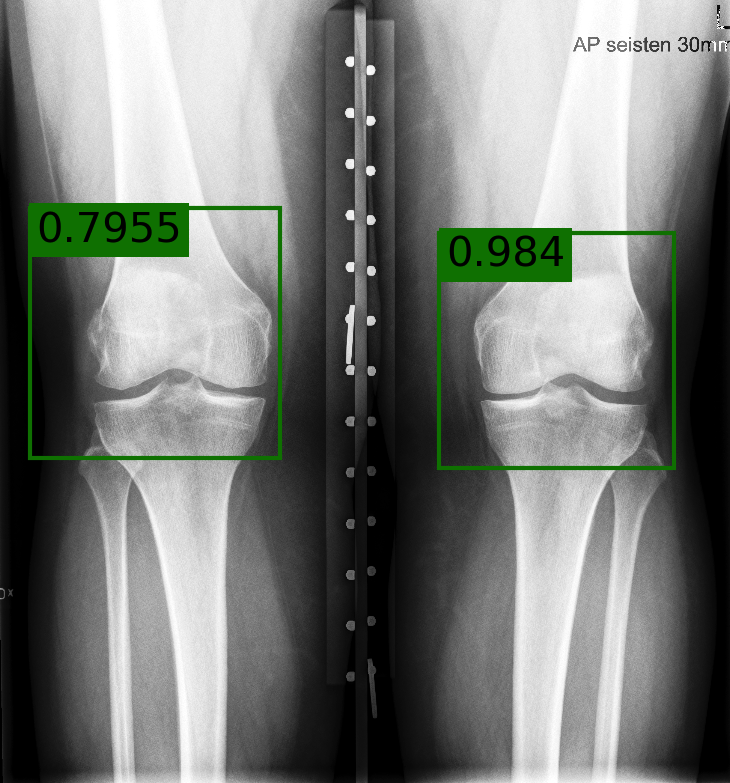}
       }
     \caption{Examples of bounding boxes produced by our method (OKOA dataset). Here, the IoU values for right and left joint are presented: a) 0.8953 and 0.77, b) 0.9373 and  0.7867, c) 0.7955 and 0.984.}
     \label{img:detex}
\end{figure}

We varied the step of displacement $p$ from 5 to 1,000 and evaluated each generated proposal by calculating IoU with the manual annotations. The best IoU score was used as a measure of quality. Eventually, we used different IoU thresholds to evaluate the best possible recall.  It can be seen from Figure \ref{img:propquality}, that on all testing datasets for each given IoU threshold our proposal algorithm reaches at least 80\% recall. Using the pre-trained SVM classifier described above, we also reached high mean IoU for all analyzed datasests (see, Table  \ref{tab:locperf}). Examples of localization are given in Figure \ref{img:detex}.

\begin{table}
\centering
\caption{Localization performance evaluation. The computations were parallelized on Intel i7 5820k CPU. Time benchmarks were averaged over 3 runs.}\label{tab:locperf}
\begin{tabular}{p{1.8cm}p{2cm}p{2.2cm}p{2.5cm}p{2.7cm}}
\hline
\noalign{\smallskip}
Dataset & \# of images & Mean IoU [\%] & Average time [s] &  Average time/image [ms] \\
\noalign{\smallskip}
\hline
\noalign{\smallskip}
MOST & 473& 0.8399 & 79.9922 &  169 \\ 
Jyv\"askyl\"a & 93 & 0.7878  & 6.4195 &  14\\
OKOA & 77& 0.7761 & 7.5264 &  16\\ 
\noalign{\smallskip}
\hline
\noalign{\smallskip}
\end{tabular}
\end{table}

\begin{figure}[!ht]
     \subfloat[Our method \label{img:ours}]{%
       \includegraphics[width=0.48\textwidth]{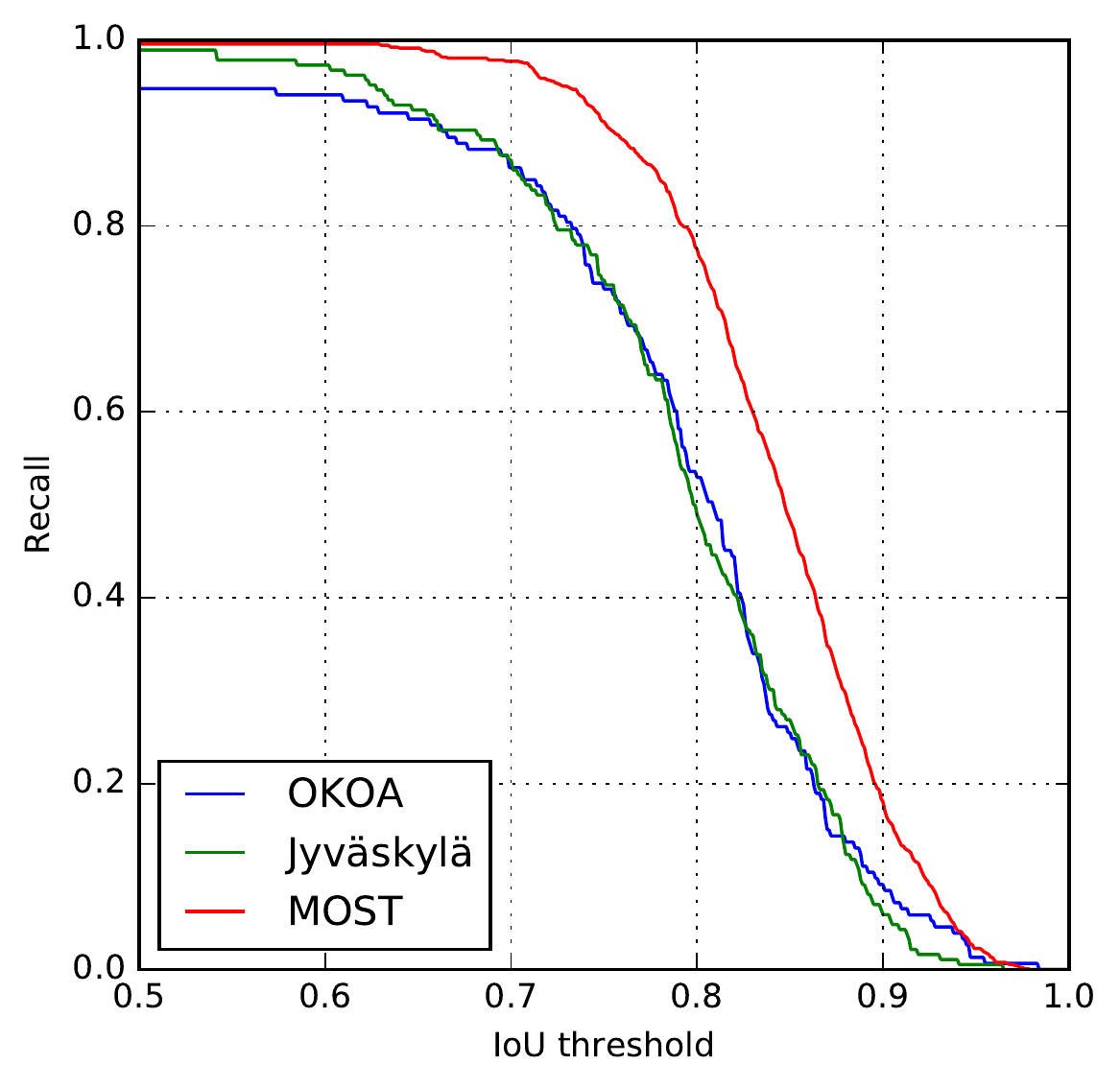}
       }
     \hfill
     \subfloat[Baseline adapted from \cite{Antony16} \label{img:baseline}]{%
       \includegraphics[width=0.48\textwidth]{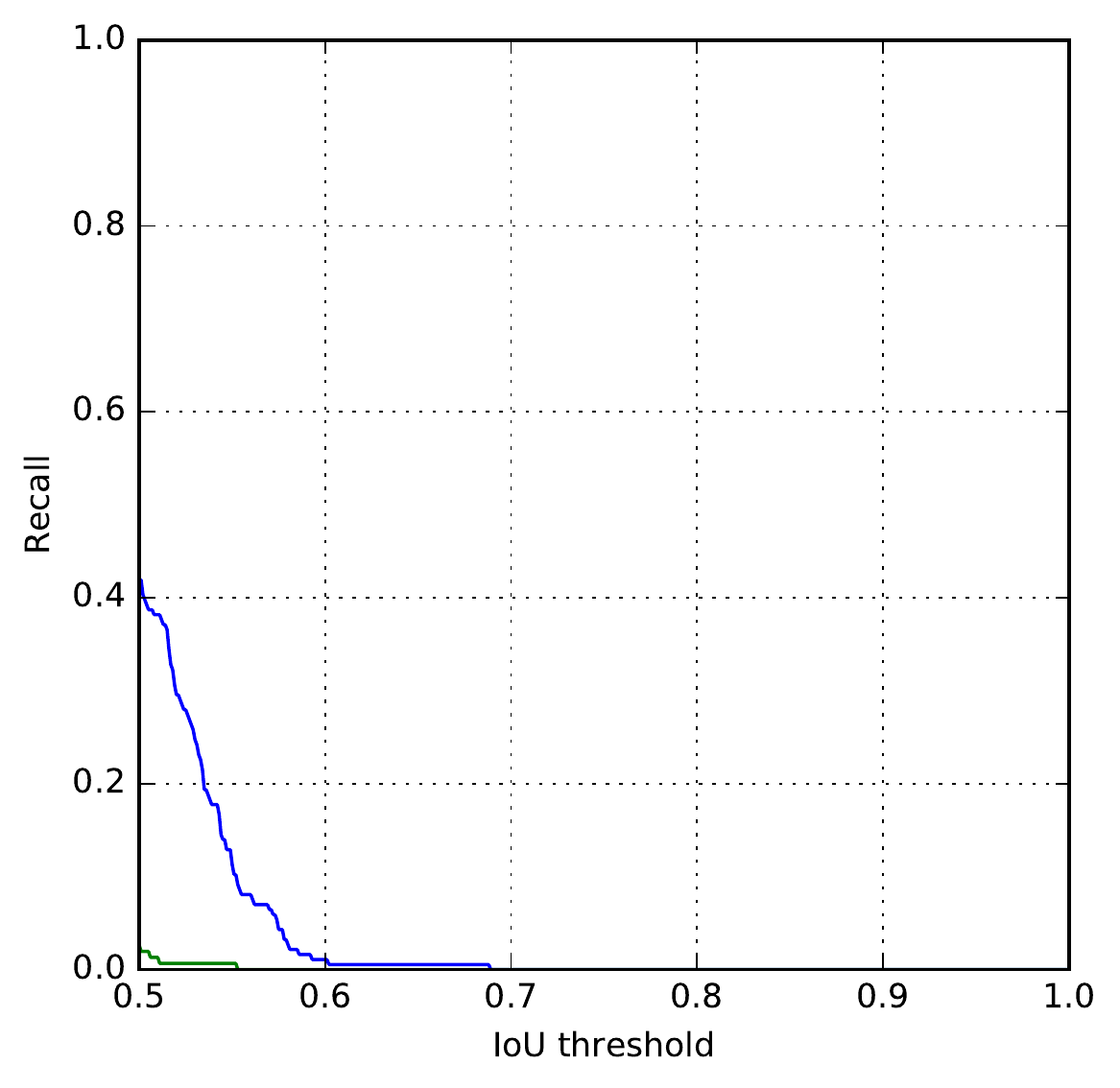}
     }
     \caption{Evaluation of recall depending on different IoU thresholds. The curves indicate a significant advantage of our method a) in comparison to the baseline b) for each of the testing datasets.}
     \label{img:baseline_comp}
\end{figure}

Apart from our own implementation, we also adopted the method described in \cite{Antony16} as a baseline (see, section \ref{related_work}). The only difference in our approach was that we used a larger window as a joint center patch  -- 40x40 pixels instead of 20x20, since with the latter, the baseline method did not perform well with our images. It can be seen from Figure \ref{img:baseline_comp}, that our method clearly outperforms the baseline on each analyzed dataset. The mean IoU values for the baseline were 0.1,0.33 and 0.26 for MOST, Jyv\"askyl\"a and OKOA datasets, respectively. 

\section{Discussion and Conclusions}
In this study, we presented a novel automatic method for knee joint localization on plain radiographs. We demonstrated our proposal generation approach, which allows avoiding using exhaustive sliding window approaches. Here, we also showed a new way to use the information about the anatomical knee joint structure for proposal generation. Moreover, we showed that the generated proposals in average are highly reliable, since a recall of above 80\% can be reached for IoU thresholds 0.5, 0.7 and 0.8. We showed that our method significantly outperforms the baseline. In the presented results, we showed that the baseline method performs comparatively similar on Jyv\"askyl\"a dataset as on OAI data in \cite{Antony16} (reported mean IoU was 0.36), however, the detector fails on MOST dataset. This can, most probably, be explained by the presence of artifacts -- parts of knee positioning frame, which are detected as joint centers.

We demonstrated, that our method generalizes well: the trained model can be used for other datasets than the one used during the training stage. Moreover, the proposed method neither requires a complex training procedure nor much computational power and memory. The developed method is mainly designed to be used for large scale knee X-ray analysis -- especially for a CAD of OA from plain knee radiographs. However, the applications are not limited to this domain only. Our approach can also be easily adapted, for example, to the hand radiographic images.

Nevertheless, some limitations of this study remain to be addressed. First of all, our results are biased due to the manual annotations -- only one person annotated the images. Secondly, the used data augmentation might include some of the false positive regions in the positive set for training the scoring block of the pipeline, which can have a negative effect on the detection. Finally, our method can be can be computationally optimized. For example, bias in the HoG-SVM block can explain a slight performance drop on Jyv\"askyl\"a and OKOA datasets. 

The method can be improved by applying the following optimizations. At first, the detection can be done on the downscaled images and then the detected joint area coordinates just need to be upscaled. We believe this optimization will significantly speed up the computations, since there is more than 10 times difference in performance between Jyv\"askyl\"a dataset and MOST. However, this might require to find new hyperparameters. The second optimization would be to reuse Sobel gradients for the overlapping ROI proposals before computing HoG features, since in our current implementation, we recomputed them for each of the proposals. Furthermore, the following post-processing step can be applied for a possible improvement of the localization performance: localization regions could be centered at the joint center -- this can be done by using a classifier from the baseline method \cite{Antony16}. However, the effect of these optimizations on the localization performance needs to be further investigated. 

To conclude, despite the limitations, our method scales with a number of cores and can also be even efficiently parallelized on GPU to achieve a high-speed detection, due to the presence of loops. For example, it can be parallelized over X and Y coordinates of joints locations. It can be calculated using the values given in Table \ref{tab:locperf}, that our data-parallel CPU implementation written in Python 2.7 already will allow to annotate more than 6,000,000 images of the size $2500\times2000$ per day (here, Jyv\"askyl\"a dataset is taken as a reference), which makes the large-scale knee X-ray studies analysis possible. Eventually, our novel approach may enable the reliable and objective knee OA CAD, which will significantly benefit OA research as well as clinical OA diagnostics. The implementation of our method will be soon released on GitHub.

\section*{Acknowledgements}
MOST is comprised of four cooperative grants (Felson – AG18820; Torner – AG18832, Lewis – AG18947, and Nevitt – AG19069) 
funded by the National Institutes of Health, a branch of the Department of Health and Human Services, 
and conducted by MOST study investigators. This manuscript was prepared using MOST data and does not necessarily reflect the opinions or views of MOST investigators.

The authors would also like to acknowledge the strategic funding of University of Oulu.
\bibliographystyle{splncs03}
\bibliography{scia2017_tiulpin} 

\end{document}